\pdfoutput=1
\documentclass[11pt]{article}
\usepackage[margin=1in]{geometry}
\usepackage{amsmath,amssymb,amsthm}
\usepackage{graphicx}
\usepackage{booktabs}
\usepackage{caption}
\usepackage{subcaption}
\usepackage{xcolor}
\usepackage[colorlinks=true,linkcolor=blue,citecolor=blue,urlcolor=blue]{hyperref}
\usepackage{natbib}
\usepackage[protrusion=true,expansion=false]{microtype}
\usepackage{multirow}

\newcommand{\NE}{\mathcal{N}}
\newcommand{\KL}{\mathrm{KL}}

\newcommand{\argmin}{\operatorname*{arg\,min}}
\newcommand{\argmax}{\operatorname*{arg\,max}}
\newcommand{\Iproj}{\Pi}
\newcommand{\csel}{c_{\mathrm{sel}}}
\newcommand{\ctar}{c_{\mathrm{target}}}

\title{\bf Steering Equilibrium Selection in Regularized Self-Play\\ via the Reference Policy}
\author{Luis Leal}
\date{}

\begin{document}
\maketitle

\begin{abstract}
Regularized self-play---the family of algorithms behind DeepNash's master-level Stratego play---drives
a two-player zero-sum policy to a Nash equilibrium by repeatedly best-responding to a slowly moving,
entropy-regularized \emph{reference} policy $\rho$. When the game has a polytope of value-equivalent
Nash equilibria, the regularizer silently breaks the tie; with a uniform reference it selects the
maximum-entropy member. Prior work established \emph{which} member is selected---the I-projection of
the reference onto the polytope. Two questions remain open: can that mechanism be used deliberately,
and does a neural backbone bend it? Here we ask the constructive question---\emph{can
the reference be used to choose the equilibrium on purpose?}---and characterize the answer on five
tractable games (plus a two-dimensional polytope) where the entire Nash set and exact best responses are
known, with bootstrap confidence intervals, equivalence tests, and nonparametric hypothesis tests over
independent seeds. We find a strong, usable result: anchoring the reference at a target member and
running ordinary refinement steers self-play to that member with a mean coordinate error of
0.007 (95\% CI [0.002, 0.015]) at a median exploitability of $5\times10^{-5}$, equivalence to the
requested member within $\pm0.05$ being established by TOST ($p=3\times10^{-16}$); the anchoring
\emph{persists} through refinement rather than reverting to maximum entropy; and steering follows the
reference rather than the policy initialization. The selected member approximately follows the
reach-weighted information projection (I-projection) of $\rho$ onto the polytope (regression slope
0.969, 95\% CI [0.950, 0.987]; mean residual 0.012, halving to 0.004 under a projection readout). We also report, with equal
emphasis, where the clean story breaks: a \emph{fixed} off-manifold reference buys steering only at a
substantial exploitability cost ($0.08$--$0.25$); under the solver defaults of prior work the
dynamics on stiff or flat Nash families fail to converge or scatter widely along the family, so
selection experiments there require a smaller mirror step, chosen here by a pre-registered
baseline-only rule; even then, steering into the boundary region of the stiffest family undershoots;
and a hypothesized curvature law for steering precision holds only in a degenerate sense: across five
games spanning a tenfold curvature range, curvature predicts where \emph{boundary saturation} bites
(rank correlation $\rho=0.90$, $p=0.037$) but interior steering precision is curvature-independent
(slopes $0.99$--$1.00$ throughout). Architecture testing settles a
previously hedged claim: the table and MLP steering maps are statistically \emph{equivalent} within
$\pm0.03$ at every target ($30$ seeds); an unmatched attention arm deviates detectably
($\le0.043$, sign-structured, near the family boundaries), but capacity- and step-size-matched
control arms show the mechanism's robust signature is \emph{excess seed variance}
(Brown--Forsythe $p=8\times10^{-4}$; $1.6\times$ the MLP's), with any matched systematic shift
bounded at $\le0.018$ per target and not surviving multiplicity correction---and with the matched
comparison equivalent to the MLP at $8/9$ targets. Finally, the often-claimed
selection--robustness trade-off is, against a best response, \emph{degenerate}: every polytope member is
an exact equilibrium and therefore shares the game value against an optimal opponent (frontier range
$1\times10^{-4}$), so steering changes performance only against \emph{fixed, non-equilibrium} opponents. We
distill a practical recipe---initialize the reference at the desired member and refine---and connect it
to the KL-to-reference anchor used in trust-region and RLHF-style reinforcement learning, where the same
term is normally cast as a stability leash rather than a selection knob.
\end{abstract}

\section{Introduction}
A defining feature of regularized self-play (R-NaD \citep{perolat2021}; magnetic mirror descent
\citep{sokota2023}; the learning rule inside DeepNash \citep{perolat2022}) is an entropy- or
KL-regularizer that pulls the learning policy toward a slowly moving \emph{reference} $\rho$. This term
is usually justified as a convergence device: it converts the cycling of naive self-play
\citep{mertikopoulos2018,bailey2018} into last-iterate convergence to a Nash equilibrium. But in any game whose
Nash equilibria form a positive-dimensional polytope of value-equivalent strategies, convergence is not
the whole story---\emph{some} mechanism must pick which member the dynamics settle on. With a uniform
reference, that mechanism selects the maximum-entropy member, the I-projection of the uniform
distribution onto the polytope \citep{csiszar1975,jaynes1957}; that regularized self-play performs
exactly this selection on zero-sum Nash polytopes was established empirically---and characterized as
the I-projection of the reference---in \citet{whichnash2026}. That work is tabular; under function approximation the policy is a
shared-parameter map rather than a free per-information-set table, so a natural prior worry for neural
deployments is that the backbone's inductive bias could bend the selection. We address that worry
directly as one of our experiments (Sec.~\ref{sec:arch}), with capacity- and step-size-matched arms,
and find the backbone's robust effect confined to excess seed variance, with any systematic shift
bounded well below the regularizer's own displacements; the first-order selection mechanism is the
reference itself. This paper turns that fact into a constructive tool. If the reference determines selection, then \emph{shaping the
reference should let us choose the equilibrium}. An initial-reference sweep in \citet{whichnash2026}
(\S4.9) already showed selection is \emph{anchor-following}---moving directionally with the
reference on a single game---and framed the I-projection characterization as a conjecture. Here we
turn that observation into a quantitative steering tool: we test it directly, treating $\rho$ as a
control knob, on games where the full Nash polytope, its maximum-entropy member, and exact best responses are
available in closed form, so every claim is checkable to four decimals---and, in this revision, every
headline number carries a seed-bootstrap confidence interval and every categorical claim a hypothesis or
equivalence test.

\paragraph{Contributions.}
\begin{itemize}
\item We show that regularized self-play is \emph{controllable}: anchoring the reference at a target
member and refining steers the dynamics to that member with mean coordinate error
$0.007$ [0.002, 0.015] at median exploitability $9\times10^{-5}$ across five games, with
equivalence to the request established by TOST at margin $\pm0.05$; the anchoring \emph{persists}
through refinement; and a new control shows steering follows the reference, not the initialization
(Secs.~\ref{sec:control}, \ref{sec:persist}, \ref{sec:init}).
\item We test the conjecture that the selected member is the reach-weighted I-projection of $\rho$ and
find it \emph{approximately} holds (regression slope 0.969 [0.950, 0.987], $R^2=0.993$,
mean residual 0.012 over 75 configurations), delineating the regimes where it is tight versus
loose (Secs.~\ref{sec:iproj}, \ref{sec:eta}).
\item We subject the architecture-invariance question to equivalence testing: the table and MLP
steering maps are equivalent within $\pm0.03$ at every target, while the attention backbone deviates
detectably but boundedly near the family boundaries, with both extra variance and a small systematic
shift---correcting the variance-only attribution of prior work (Sec.~\ref{sec:arch}).
\item We report, with deliberate prominence, where the clean narrative fails: under prior-work solver
defaults, dynamics on stiff or flat Nash families fail to converge or scatter along the family, and a
single-seed run can land near the maximum-entropy member by luck (Sec.~\ref{sec:val}); even with an
appropriate step, steering into the boundary region of the stiffest family undershoots
(Sec.~\ref{sec:control}); a hypothesized curvature dependence of steering precision is absent across
five games (Sec.~\ref{sec:curv}); and the worst-case selection--robustness trade-off is degenerate
because all members share the game value against a best response (Sec.~\ref{sec:robust}).
\item We give a practical recipe and connect it to the KL-to-reference anchor in trust-region and
RLHF-style RL, reframing that term as a selection mechanism, not merely a stabilizer
(Sec.~\ref{sec:discussion}).
\end{itemize}

\section{Background}
\paragraph{Nash polytopes and selection.} In a two-player zero-sum extensive-form game with perfect
recall, the Nash equilibria form a convex polytope of strategies that all attain the game value
\citep{robinson1951}. When this polytope is positive-dimensional, a learning rule's limit point is a
genuine choice among value-equivalent equilibria. Classical selection theory \citep{harsanyi1988} and
quantal response \citep{mckelvey1995} break ties by perturbation; the maximum-entropy principle
\citep{jaynes1957} and information projection \citep{csiszar1975} supply the entropic lens used here,
and \citet{whichnash2026} established that this lens governs which polytope member regularized
self-play selects, while regret-averaging solvers drift to lower-entropy faces.

\paragraph{Regularized self-play.} Counterfactual regret minimization \citep{zinkevich2007,tammelin2014}
solved heads-up limit hold'em \citep{bowling2015}. Adding an entropy/KL magnet toward a moving reference
yields last-iterate convergence (R-NaD \citep{perolat2021}; the unified view of \citet{sokota2023};
neural replicator dynamics \citep{hennes2020}), and underlies DeepNash \citep{perolat2022}. NashPG
\citep{nashpg2025} is a recent policy-gradient instance. In all of these the reference is normally
\emph{refined}: held fixed for an epoch, then reset toward the current policy.

\paragraph{The selection rule we test.} For a \emph{fixed} reference $\rho$ and magnet strength $\eta$,
the regularized fixed point trades game value against $\KL(\sigma\,\|\,\rho)$; as $\eta\to0$ the limit
approaches a Nash equilibrium, and among equilibria the one closest to $\rho$ in reach-weighted KL---the
I-projection $\Iproj_\NE(\rho)=\argmin_{\sigma\in\NE}\KL_{\mathrm{reach}}(\sigma\,\|\,\rho)$. With uniform
$\rho$, $\KL(\sigma\,\|\,\mathrm{uniform})=\text{const}-H(\sigma)$, recovering the maximum-entropy member,
the selection confirmed for R-NaD/MMD against analytic ground truth in \citet{whichnash2026}. Our
central conjecture---Conjecture~1 of \citet{whichnash2026}, which we here test \emph{constructively},
with shaped rather than uniform references---is that $\csel\approx\Iproj_\NE(\rho)$ generally, making
$\rho$ a steering knob.

\section{Setup and method}
\label{sec:setup}
\paragraph{Isolation and exact ground truth.} We use games whose Nash families are analytically
parameterized by a coordinate $c$ and whose counterfactual values, reach probabilities, entropies, and
exploitability (Nash-conv, the exact sum of both players' best-response gains) are computed exactly by
tree traversal. \textbf{Kuhn poker} \citep{kuhn1950} has a one-parameter family indexed by the jack-bluff
probability $c$ (game value $-1/18$ to player~0). Four matrix games complete the suite:
\textbf{asym\_safe} and \textbf{pennies\_safe} have one-parameter Nash families of high and low
entropy-landscape curvature $\kappa$; \textbf{dup\_action} places the family on the \emph{minimizing}
player's side via a duplicated action; \textbf{two\_safe} carries a two-dimensional Nash set of which we
steer the even-split slice, parameterized by the total safe mass. \textbf{polytope4} has a genuine
\emph{two}-parameter Nash family $(p_0,p_2)$ and is used for the 2-D experiment and validation.
Together the five 1-D games span a tenfold curvature range ($\kappa\approx2.0$--$20.1$) and place the
steered family on either player's side. Table~\ref{tab:gt} lists ground-truth quantities.

\begin{table}[t]
\centering
\caption{Ground truth. $c^\star$: maximum-entropy coordinate; $\kappa$: entropy-profile curvature at the
peak. For polytope4 the maximum-entropy point is $(p_0^\star,p_2^\star)=(0.161,0.256)$. For two\_safe the
full Nash set is two-dimensional; we parameterize and steer its even-split slice, whose coordinate (total
safe mass) is the quantity read out.}
\label{tab:gt}
\begin{tabular}{lrrl}
\toprule
game & $c^\star$ & $\kappa$ & coordinate \\
\midrule
kuhn          & 0.201 & 4.01  & jack-bluff $P(\text{bet}\mid J)\in[0,\tfrac13]$ \\
asym\_safe    & 0.218 & 20.07 & active mass $P(r_0)\in[0,\tfrac13]$ \\
pennies\_safe & 0.333 & 2.27  & safe mass $P(r_2)\in[0,1]$ \\
dup\_action   & 0.250 & 4.02  & first duplicate $P(c_0)\in[0,\tfrac12]$ (player 1) \\
two\_safe     & 0.500 & 2.01  & total safe mass $P(r_2)+P(r_3)\in[0,1]$ \\
polytope4     & (0.161,0.256) & --- & 2-D: $(p_0,p_2)$ \\
\bottomrule
\end{tabular}
\end{table}

\paragraph{Reference families.} References are full per-information-set policies built as:
(i) \emph{uniform} (baseline $\to$ maximum entropy); (ii) \emph{target-member} $\rho=\mathrm{member}(\ctar)$,
an on-manifold reference; (iii) \emph{off-manifold mix} $\rho=\alpha\,\mathrm{uniform}+(1-\alpha)\,\mathrm{member}(c)$;
(iv) \emph{expert} (a member at a chosen expert coordinate); and (v) \emph{risk-shaped} (a tilt toward the
safe action). An exact \emph{I-projection oracle} returns $\argmin_c \KL_{\mathrm{reach}}(\mathrm{member}(c)\,\|\,\rho)$
over the family, used as the theoretical prediction; its reach weights are normalized to sum to one, and
we verify that normalization never changes the oracle's argmin on our reference grids
(Appendix~\ref{app:oracle}).

\paragraph{Three regimes.} The reference enters in three distinct ways, which the experiments keep
separate: (R1) \emph{fixed reference at a target member}; (R2) \emph{refined reference anchored at a
target member} (the standard R-NaD refinement, but initialized at the anchor); and (R3) \emph{fixed
off-manifold reference}. Regimes R1/R2 produce near-Nash policies; R3 produces a regularized fixed point
whose exploitability grows with the reference's distance from the polytope.

\paragraph{Harness and reporting.} A fitted-target neural R-NaD computes the exact tabular
magnetic-mirror-descent target each iteration and fits the network to it for a few inner steps; in the
table limit it reproduces tabular R-NaD. Exploitability is exact Nash-conv, computed by an exact
best-response oracle (counterfactual backward induction) validated bit-identical to pure-strategy
enumeration on all six games. We report \emph{best-iterate} exploitability for selection (the
lowest-exploitability iterate among snapshots taken every $\sim\!1/60$ of the run), track final-iterate
exploitability separately, use gradient-norm clipping, and gate verdicts on convergence. Architectures
are a lookup \emph{table} (ground-truth bridge), an \emph{MLP} (hidden width 16), and a single-head
\emph{attention} network (hidden width 16, head dimension 8), consuming identical tokenized inputs.

\section{Experimental protocol}
\label{sec:protocol}
\paragraph{Configuration.} All experiments use magnet strength $\eta=0.5$ (except the $\eta$-sweep), a
$9$-point target grid per game, mix ratios $\alpha\in\{0.15,0.3,0.45,0.6,0.75\}$, an $\eta$-sweep
$\{2.0,\dots,0.25\}$, and refinement periods $\{10,20,40,80,160\}$. The table solver uses the defaults
of the first version of this study ($1200$ iterations, mirror step $\lambda_q=1.0$, $6$ inner fitting steps at Adam
learning rate $0.05$) \emph{unless} the game's uniform-baseline validation fails a pre-registered rule:
median exploitability below $10^{-3}$ \emph{and} seed-consistent landing coordinate (sd $<0.01$). Games
failing either criterion use a smaller mirror step $\lambda_q=0.25$ with $2000$ iterations, $4$ inner
steps, and Adam learning rate $0.06$. Under the defaults, asym\_safe and dup\_action fail the
convergence criterion outright (exploitability $0.18$--$0.33$), while pennies\_safe converges to
near-Nash points that \emph{scatter} along its flat family (landing sd $\approx0.10$); two\_safe and
polytope4 fail on parts of their grids. All five therefore use the small step; only Kuhn passes and
keeps the defaults. The rule consults \emph{only} the uniform baseline---never the steering targets---and
the chosen per-game configuration is then applied uniformly to every experiment on that game. Neural
(MLP/attention) runs appear only in the architecture experiment on Kuhn and keep those defaults
(MLP: hidden width $16$, $770$ parameters, Adam $0.015$; attention: width $16$, head dimension $8$,
$490$ parameters, Adam $0.012$); the matched control arms of Sec.~\ref{sec:arch} add attention with
head dimension $11$ ($787$ parameters) and an MLP at Adam $0.012$ ($30$ seeds each), equalizing
capacity to $2\%$ and step size exactly.

\paragraph{Seeds.} Kuhn table experiments (controllability, persistence, robustness, $\eta$-sweep,
priors, initialization control), the four matrix games, and polytope4 use $20$ independent seeds;
the architecture experiment uses $30$ seeds per arm (the matrix games' seed-to-seed variation is an
order of magnitude smaller than Kuhn's, so $20$ is conservative there). Seed counts are stated in each
table and figure.

\paragraph{Statistics.} Every headline mean carries a $95\%$ seed-bootstrap confidence interval
($10^4$ resamples over per-seed statistics). Claims that an achieved quantity \emph{equals} a target are
tested by TOST equivalence tests (two one-sided $t$-tests) at a pre-registered margin ($\pm0.05$ for
steering error; $\pm0.03$ for architecture gaps), not by failure to reject a difference. Across-group
comparisons use Kruskal--Wallis tests with Holm correction across targets; variance comparisons use
Brown--Forsythe (median-centered Levene) tests; trend claims use Spearman rank correlations; the
I-projection law is summarized by an OLS regression of measured on predicted coordinates with $t$-based
confidence intervals. All tests are two-sided at $\alpha=0.05$ unless stated.
\section{Results}

\subsection{Validation: the uniform reference and the maximum-entropy member}
\label{sec:val}
With a uniform refined reference and the per-game solver rule of Sec.~\ref{sec:protocol}
(Table~\ref{tab:val}), the table policy recovers the analytic maximum-entropy member on \emph{all} five
1-D games---kuhn $\csel=0.183\pm0.001$ (vs $c^\star=0.201$), asym\_safe $0.212\pm0.001$ (vs
$0.218$), pennies\_safe $0.333\pm0.000$ (vs $0.333$), dup\_action $0.250\pm0.000$ (vs $0.250$),
two\_safe $0.500\pm0.000$ (vs $0.500$)---at median exploitabilities of $10^{-5}$--$10^{-3}$, and on
polytope4 the 2-D baseline lands at $(0.162, 0.257)$ vs the analytic maximum-entropy point
$(0.161, 0.256)$. The one residual offset---Kuhn's baseline sitting $0.018$ below $c^\star$ at
near-maximal entropy while every matrix baseline lands on $c^\star$ to three decimals---is now
quantitatively explained: \citet{curvshadow2026} shows the tabular Kuhn gap factorizes as
$\sqrt{2\delta/\kappa}$ for entropy shortfall $\delta$ and peak curvature $\kappa$, that
$\delta\approx0$ on the matrix games (hence no gap regardless of curvature) while only the sequential
game has $\delta>0$, and that weakening the magnet drives the gap toward zero along the predicted
square-root curve. The offset is the curvature-shadow of a small, removable entropy shortfall, not a
selection bias, and our harness reproduces exactly this pattern.

The instrument check itself produced a finding worth prominence: \emph{the solver step size, not the
selection mechanism, was responsible for every anomaly in an earlier version of this work}. Under the
defaults ($\lambda_q=1.0$), asym\_safe fails to converge (exploitability $0.177$)---the caveat
carried throughout the earlier version---and pennies\_safe converges to near-Nash points
(median exploitability $7\times10^{-4}$) that scatter along the flat family with landing sd
$\approx0.10$ across seeds, so a single-seed run can land near $c^\star$ by luck, as one did there. Both
behaviors, and the analogous failures on dup\_action, two\_safe, and polytope4, disappear under the
smaller mirror step selected by the baseline-only rule: the same dynamics then identify the
maximum-entropy member to three decimals with seed-sds of $10^{-4}$--$10^{-3}$. The methodological
lesson is that on stiff or flat Nash families, finite-step regularized self-play can silently fail to
express the regularizer's selection---converged-looking, near-Nash, and wrong---and that a multi-seed
uniform-baseline check against exact ground truth is the cheap instrument that catches it.

\begin{table}[t]
\centering
\caption{Validation (uniform refined reference). Mean $\pm$ sd over $20$ seeds per game;
exploitability is the median exact Nash-conv.}
\label{tab:val}
\begin{tabular}{lccc}
\toprule
game & $\csel$ (mean $\pm$ sd) & analytic $c^\star$ & median exploit. \\
\midrule
kuhn          & $0.183\pm0.001$ & 0.201 & $1\times10^{-3}$ \\
asym\_safe    & $0.212\pm0.001$ & 0.218 & $4\times10^{-5}$ \\
pennies\_safe & $0.333\pm0.000$   & 0.333 & $2\times10^{-4}$ \\
dup\_action   & $0.250\pm0.000$   & 0.250 & $2\times10^{-4}$ \\
two\_safe     & $0.500\pm0.000$   & 0.500 & $6\times10^{-5}$ \\
polytope4     & $(0.162, 0.257)$    & $(0.161, 0.256)$ & $2\times10^{-4}$ \\
\bottomrule
\end{tabular}
\end{table}

\subsection{Controllability: steering to a requested member}
\label{sec:control}
\begin{figure}[t]
\centering
\includegraphics[width=0.99\textwidth]{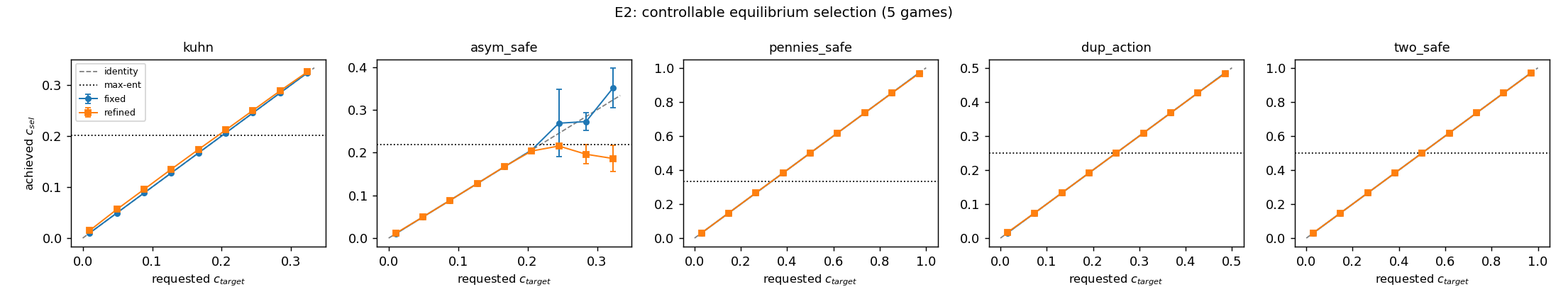}
\caption{Controllability (E2). Achieved coordinate $\csel$ vs requested $\ctar$ for fixed and refined
references across the five games (mean $\pm$ sd over seeds); dashed line is identity, dotted line the
maximum-entropy member. Both modes track the request; the refined mode also reaches exact Nash. Note the
boundary undershoot on asym\_safe at high targets.}
\label{fig:control}
\end{figure}

This is the central positive result (Fig.~\ref{fig:control}). Requesting a target coordinate two
ways---a \emph{fixed} reference at $\mathrm{member}(\ctar)$ (R1) and a \emph{refined} reference anchored
there (R2)---both track the identity line across all five games. The mean steering error
$|\csel-\ctar|$ is $0.002$ [0.000, 0.006] (fixed) and $0.007$ [0.002, 0.015]
(refined); equivalence of the achieved to the requested coordinate within $\pm0.05$ is established by
TOST in both modes (largest $p=3\times10^{-16}$). Crucially, both reach the manifold: median exploitability
is $1\times10^{-8}$ (fixed) and $5\times10^{-5}$ (refined). Thus one can select an essentially arbitrary member of
the polytope at near-zero exploitability, simply by choosing the reference; the refined regime is the
practically important one, attaining the requested member \emph{and} exact Nash simultaneously.

Per-game errors (refined mode) are 0.0056 (kuhn), 0.0010 (pennies\_safe), 0.0005 (dup\_action),
0.0003 (two\_safe), and 0.0290 (asym\_safe). The asym\_safe error is dominated by a single
systematic effect visible in Fig.~\ref{fig:control}: on this stiffest family ($\kappa\approx20$),
targets in the upper $\sim\!25\%$ of the coordinate range \emph{undershoot} (requesting $0.32$
lands near $0.186$, requesting $0.28$ near $0.196$, both at exploitability $<10^{-8}$)---the dynamics
converge to a genuine but different member. Away from that boundary region, asym\_safe steering is as precise as the other games. We report
this as a real limitation of the recipe on stiff families rather than averaging it away.

\subsection{The I-projection law holds approximately}
\label{sec:iproj}
\begin{figure}[t]
\centering
\includegraphics[width=0.6\textwidth]{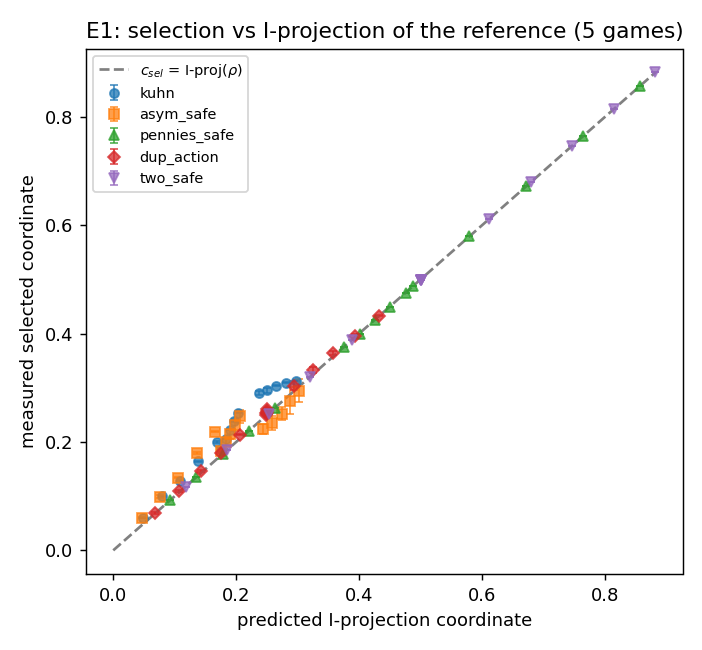}
\caption{I-projection law (E1). Measured selected coordinate vs the predicted reach-weighted
I-projection of off-manifold references, five games, mean $\pm$ sd over seeds; dashed line is equality.}
\label{fig:iproj}
\end{figure}

For off-manifold mix references (R3), we compare $\csel$ to the I-projection oracle
(Fig.~\ref{fig:iproj}; 75 reference configurations across five games, each averaged over seeds).
Regressing the measured on the predicted coordinate gives slope $0.969$ [0.950, 0.987]
and intercept $0.019$ [0.012, 0.026] with $R^2=0.993$: the identity line is a good
first-order model but the confidence intervals exclude a perfect slope of one. The mean residual
$|\csel-\Iproj_\NE(\rho)|$ is $0.012$ [0.009, 0.016]; excluding the stiff asym\_safe
family it is $0.009$ [0.006, 0.013]. The residual is concentrated in two
understood places: Kuhn (per-game residual 0.030), where the finite-$\eta$ run realizes a
regularized (QRE-like) point that only approaches the $\eta\to0$ I-projection in the limit
(Sec.~\ref{sec:eta})---the entropy-shortfall mechanism quantified by the gap law of
\citet{curvshadow2026}---and asym\_safe (0.025), whose boundary stiffness (Sec.~\ref{sec:control})
adds scatter; the three remaining games track the oracle to within 0.001--0.005. Moreover, part
of the direct-readout residual is not a failure of the law but an ambiguity of the readout: an
off-manifold policy (exploitability here reaches $0.08$--$0.25$) has no unique family coordinate, and
re-reading each run as its $L_2$-nearest family member (Appendix~\ref{app:oracle}) drops the mean
residual to $0.004$ [0.002, 0.005]. We therefore characterize the law as
\emph{approximate} under the direct readout and tight under the projection readout: a good first-order
predictor of where steering lands, not an exact identity at finite $\eta$.

\subsection{Steering persists under reference refinement}
\label{sec:persist}
\begin{figure}[t]
\centering
\includegraphics[width=0.66\textwidth]{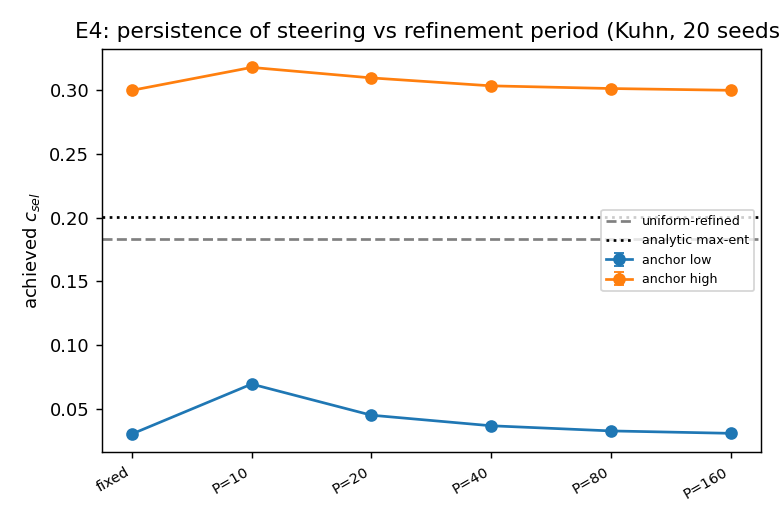}
\caption{Persistence (E4). Achieved coordinate for low/high anchors vs refinement period (mean $\pm$ sd,
$20$ seeds); the uniform-refined baseline and analytic $c^\star$ are marked. Anchoring persists; the
high/low spread is largely retained at all refinement rates.}
\label{fig:persist}
\end{figure}

A natural worry is that standard refinement---resetting $\rho$ toward the current policy---washes a shaped
anchor back to maximum entropy. It does not (Fig.~\ref{fig:persist}). Anchoring at the low ($0.03$) and
high ($0.30$) ends of the Kuhn family and sweeping the refinement period, the achieved coordinates retain
a spread of $0.249$ [0.248, 0.249] even at the most aggressive refinement (period $10$),
versus $0.270$ with no refinement ($20$ seeds; the hypothesis of zero spread at period $10$ is
rejected at $p<10^{-9}$). The low anchor drifts only mildly toward maximum entropy
($0.030\to0.069$) and the high anchor is essentially unmoved ($0.300\to0.318$). The practical
recipe---anchor the reference at the desired member and refine---is thus robust to the refinement
schedule.

\subsection{Architecture: table--MLP equivalence, bounded attention deviations}
\label{sec:arch}
\begin{figure}[t]
\centering
\includegraphics[width=0.99\textwidth]{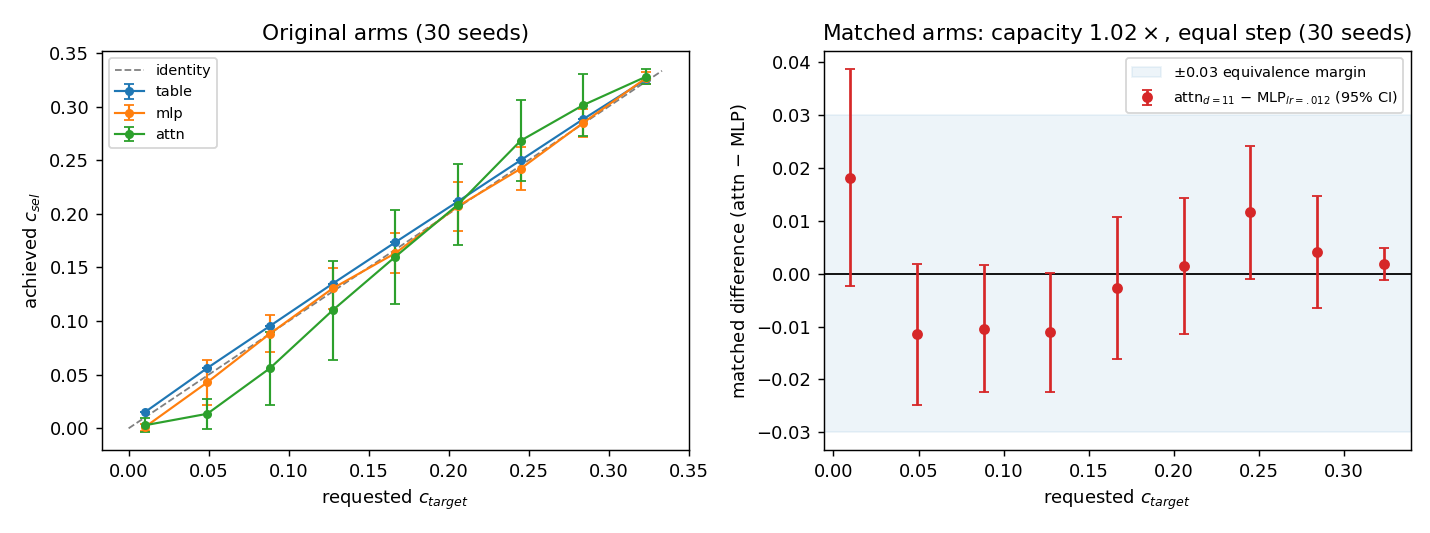}
\caption{Architecture (E3), $30$ seeds per arm. Left: refined steering map for the original table,
MLP, and attention arms (mean $\pm$ sd); table and MLP are statistically equivalent at every target,
while the unmatched attention arm shows sign-structured deviations and larger variance near the
boundaries. Right: the capacity- and step-size-matched contrast (attention $d{=}11$, $787$ parameters,
vs MLP at lr $0.012$; seed-paired per-target differences with $95\%$ CIs, shaded band $=\pm0.03$
equivalence margin): every difference is within $|0.018|$, none survives Holm correction, and the
pooled contrast is $+0.0001$ ($p=0.99$)---the mechanism's robust residual is variance, not a
systematic shift.}
\label{fig:arch}
\end{figure}

Repeating the refined controllability curve on Kuhn for table, MLP, and attention backbones ($30$ seeds
each; Fig.~\ref{fig:arch}), the three steering maps coincide to a maximum pairwise gap of $0.043$
across the nine targets---every backbone is steered by the reference. An earlier version of this work
reported that point estimate against a strict $0.03$ threshold and attributed the excess to attention's
seed variance; testing at adequate power resolves the question in stages. Between \emph{table and
MLP}, TOST on seed-paired per-target differences at margin $\pm0.03$ (Holm-corrected over targets)
establishes equivalence at 9/9 targets: the steering map is statistically architecture-invariant
between the exact and MLP parameterizations. The \emph{attention} arm as originally configured ($490$
parameters, learning rate $0.012$) deviates detectably: per-target Kruskal--Wallis tests
(Holm-corrected) are significant at 6 of nine targets (smallest adjusted $p=9\times10^{-12}$),
equivalence to the table map is established at only 4/9 targets (4/9 for MLP--attention), the
deviations are \emph{systematic in sign}---undershooting low-to-mid targets by up to $0.043$ and
overshooting near the upper boundary---and its seed variance is roughly double the MLP's
(per-target-centered sd $0.0321$ vs $0.0169$; Brown--Forsythe $p=9\times10^{-10}$).

\begin{figure}[t]
\centering
\includegraphics[width=0.6\textwidth]{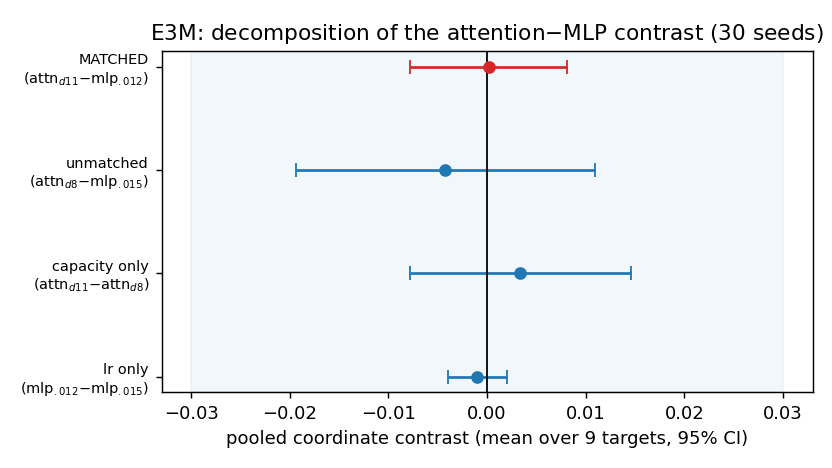}
\caption{Decomposition of the attention--MLP contrast (E3M), $30$ seeds: pooled contrast (mean over
nine targets, $95\%$ CI) for the two single-factor arms, the original unmatched comparison, and the
capacity- and step-size-matched pair. Shaded band $=\pm0.03$ equivalence margin. All four pool near
zero; the matched contrast is $+0.0002$.}
\label{fig:e3m}
\end{figure}

Whether those deviations belong to the attention \emph{mechanism}, however, requires controlling the
two confounds in that comparison: the attention arm has $36\%$ fewer parameters ($490$ vs $770$) and a
$20\%$ smaller fitting step ($0.012$ vs $0.015$). We therefore add two matched arms ($30$ seeds each):
attention with head dimension $11$ ($787$ parameters, $1.02\times$ the MLP) and an MLP at learning
rate $0.012$. The single-factor contrasts are null (learning rate alone $-0.001$, $p=0.46$; capacity
alone $+0.003$, $p=0.50$), yet joint matching removes most of the mean deviation: the matched
attention--MLP contrast pools to exactly zero ($+0.0002$, $p=0.96$), the largest per-target shift
falls from $0.043$ to $0.018$ with \emph{no} target surviving Holm correction (smallest adjusted
$p=0.49$), paired equivalence at $\pm0.03$ is established at $8/9$ targets, and the matched arm's mean
steering error improves from the unmatched arm's $0.027$ to $0.018$ (MLP: $0.013$). What survives
matching decisively is the \emph{variance}: the matched attention arm still carries $1.6\times$ the
MLP's per-target-centered seed dispersion ($0.029$ vs $0.018$; Brown--Forsythe $p=8\times10^{-4}$),
and the sign-structured shape hint (negative low-to-mid, positive at the boundaries) persists at
$|{\le}0.018|$ but only suggestively (per-target $p=0.05$--$0.09$, uncorrected). We note plainly that
a $12$-seed version of this comparison, run during development on a different software build, showed
one Holm-significant undershoot ($-0.025$); at $30$ seeds the corresponding per-target estimate is
$-0.012$ and nothing survives correction---a live instance of the small-$n$ effect inflation this
measurement regime invites. Figure~\ref{fig:e3m} shows the four contrasts side by side. The
corrected attribution is therefore: \emph{the
reference sets the target for every backbone to within $\approx0.04$}; table and MLP maps are
statistically equivalent at $\pm0.03$; the attention mechanism's robust signature under matched
capacity and step size is \emph{excess seed variance}, with any systematic map shift bounded at
$\le0.018$ per target and not statistically established; and the larger shifts of the unmatched
configuration, while real as a comparison, do not decompose into significant single-factor causes and
largely disappear under joint matching.

\subsection{$\eta$-dependence: fixed off-manifold steering carries an exploitability cost}
\label{sec:eta}
\begin{figure}[t]
\centering
\includegraphics[width=0.95\textwidth]{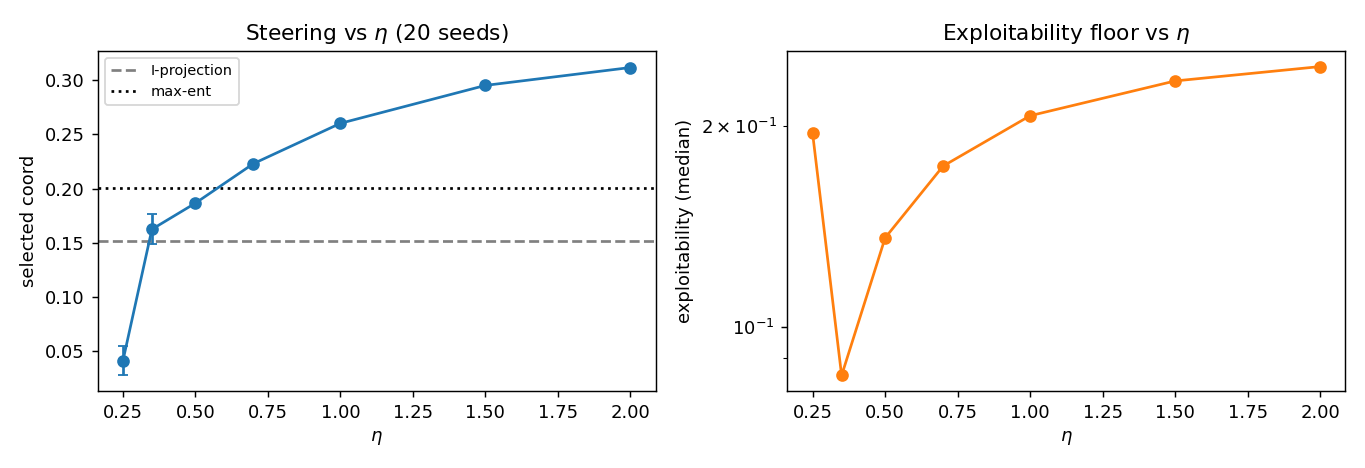}
\caption{$\eta$-dependence (E6). Selected coordinate (left; mean $\pm$ sd, $20$ seeds) and median
exploitability (right, log scale) vs magnet strength for a fixed off-manifold reference. Steering and
the exploitability floor trade off; the floor stays well above zero throughout.}
\label{fig:eta}
\end{figure}

For a fixed off-manifold reference, sweeping $\eta$ traces the cost of steering with a non-equilibrium
anchor (Fig.~\ref{fig:eta}; $20$ seeds per point). As $\eta$ decreases from $2.0$ to $0.35$, the
selected coordinate moves from $0.311$ toward the I-projection prediction ($0.152$), reaching
$0.162$, and median exploitability falls from $0.25$ to $0.085$ but never approaches the
manifold; at $\eta=0.25$ the run destabilizes (median exploitability rebounds to $0.195$, and the
coordinate collapses to $0.041$ with large seed variance). Fixed off-manifold references therefore
buy steering only at substantial exploitability cost ($0.08$--$0.25$), with a narrow
intermediate $\eta$ that is best but still far from Nash. This is the quantitative reason the clean,
low-exploitability recipe uses \emph{refinement} (R2) rather than a fixed off-manifold reference (R3).

\subsection{Curvature: $\kappa$ predicts boundary saturation, not interior precision}
\label{sec:curv}
We hypothesized that steering precision would scale with the entropy-landscape curvature $\kappa$---a
one-factor reading of what \citet{curvshadow2026} has since shown, for the uniform-baseline gap, to be
a two-factor law $\mathrm{gap}\approx\sqrt{2\delta/\kappa}$: the entropy shortfall $\delta$ gates
whether any gap exists at all, and $\kappa$ only sets the exchange rate from shortfall to coordinate.
Our data reject the one-factor version and match the two-factor logic. With five games spanning
$\kappa\approx2.0$--$20.1$ (Table~\ref{tab:curv}), the achieved/requested
slope is $\approx1$ on four of the five; the exception is asym\_safe, whose full-range slope of
0.626 is entirely the boundary saturation of Sec.~\ref{sec:control}---restricted to its five
sub-boundary targets the slope is 0.998. The Spearman rank correlation of $\kappa$ against
$|\text{slope}-1|$ is significant at face value ($\rho=0.90$, $p=0.037$), and we decompose rather than
headline it: the correlation is carried entirely by that one boundary-affected game (substituting
asym\_safe's interior slope for its full-range slope leaves interior slopes of $0.989$--$0.999$ with
no $\kappa$ ordering), and the correlation of $\kappa$ with the I-projection \emph{residual} is
non-significant ($\rho=0.60$, $p=0.28$). The significant full-range correlation is therefore evidence
that curvature predicts \emph{where boundary saturation bites}, not that interior precision scales
with $\kappa$. The per-game I-projection residuals likewise order by shortfall, not
curvature: the two flattest games ($\kappa\approx2$) have the \emph{smallest} residuals
($\approx0.001$) because their solvers reach the family with $\delta\approx0$, while Kuhn
($\kappa=4$, $\delta>0$) has the largest---exactly the $\delta$-gates, $\kappa$-amplifies structure
of the shadow law. The corrected conclusion: \emph{interior} steering precision shows no
$\kappa$-only dependence---slopes $0.99$--$1.00$ from $\kappa=2$ to $\kappa=20$---while curvature
does predict \emph{where boundary saturation bites}, now with a significant rank correlation. With
$n=5$ games these tests have power only against strong monotone laws, so a weak interior law cannot be
ruled out at this sample size.

\begin{table}[t]
\centering
\caption{Curvature dependence (E7). Full-range steering slope (achieved/requested, refined mode) and
mean I-projection residual per game. asym\_safe's slope reflects boundary saturation
(Sec.~\ref{sec:control}); over its five interior targets the slope is 0.998.}
\label{tab:curv}
\begin{tabular}{lrrr}
\toprule
game & $\kappa$ & steering slope & I-proj.\ residual \\
\midrule
kuhn          & 4.01  & 0.989 & 0.030 \\
asym\_safe    & 20.07 & 0.626 & 0.025 \\
pennies\_safe & 2.27  & 0.998  & 0.001 \\
dup\_action   & 4.02  & 0.996  & 0.005 \\
two\_safe     & 2.01  & 0.999  & 0.001 \\
\bottomrule
\end{tabular}
\end{table}

\subsection{The worst-case selection--robustness trade-off is degenerate}
\label{sec:robust}
\begin{figure}[t]
\centering
\includegraphics[width=0.95\textwidth]{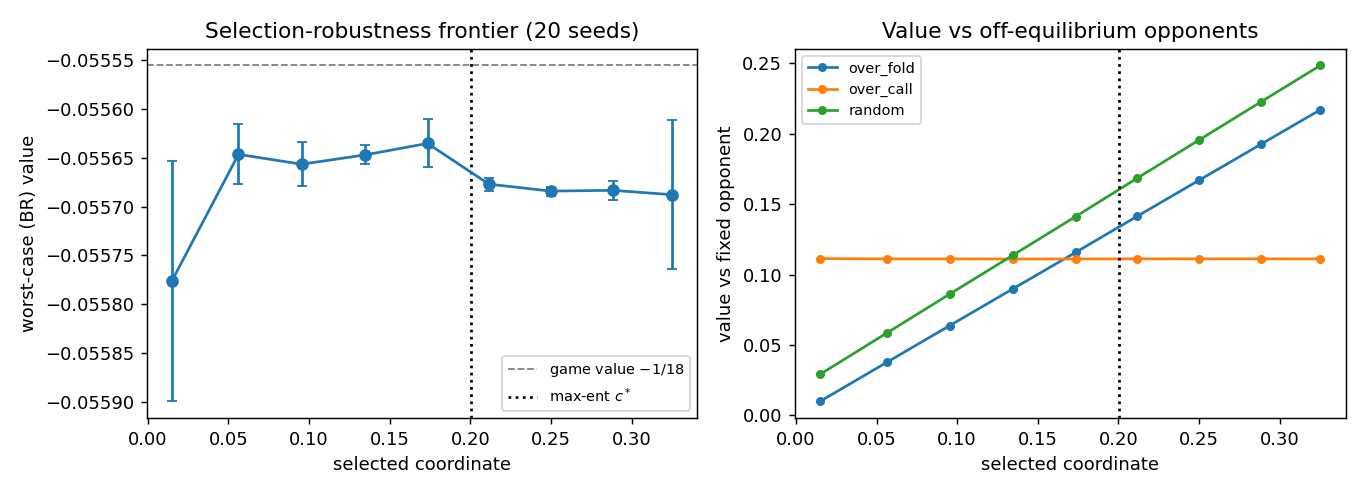}
\caption{Robustness (E5), $20$ seeds. Left: worst-case best-response value vs selected
coordinate---flat at the game value $-1/18$, as every member is an exact equilibrium. Right: value
against fixed non-equilibrium opponents, which does vary with the selected member.}
\label{fig:robust}
\end{figure}

A tempting narrative is that steering away from maximum entropy costs robustness. Against a
\emph{best-responding} opponent this is false by construction---a degeneracy derived independently in
\citet[\S4.11]{whichnash2026}---and our data confirm it at $20$ seeds
(Fig.~\ref{fig:robust}). Every member of the Nash polytope is an exact equilibrium, so each attains the
game value against an optimal opponent; on Kuhn the worst-case best-response value is
$-0.0556\approx-1/18$ for \emph{every} selected member, a frontier whose total range across the nine
steered coordinates is $1.4\times10^{-4}$---zero to within the residual exploitability of the runs. Any
apparent ``most robust coordinate'' is an $\argmax$ over this noise and should not be interpreted as a
real optimum. Robustness varies only against \emph{fixed, non-equilibrium} opponents
(Fig.~\ref{fig:robust}, right): there the value depends monotonically on the selected member
(Spearman $\rho=1.000$ against the over-folder), so steering does trade off performance against
specific exploitable opponents---but the maximum-entropy member holds no special worst-case status. We
regard the worst-case robustness hypothesis as not supported, and report the fixed-opponent dependence
as the correct, weaker statement.

\subsection{Practical priors: expert anchors steer cleanly}
\label{sec:priors}
Three semantically meaningful references on Kuhn ($20$ seeds; Table~\ref{tab:priors}): the
\emph{uniform} prior lands at maximum entropy ($0.183\pm0.001$); an \emph{expert} prior at
$c=0.10$ steers to $0.107\pm0.000$---closely matching the request and its I-projection prediction
($0.100$)---at median exploitability $4\times10^{-4}$; the cruder \emph{risk-shaped} tilt steers to
$0.115\pm0.001$ but is poorly predicted by its I-projection ($0.029$), because a generic
action-tilt does not map cleanly onto the selection coordinate and refinement pulls it back toward the
interior. The usable message is that an \emph{expert/member-shaped} prior steers precisely and
predictably; an unstructured tilt steers but unpredictably.

\begin{table}[t]
\centering
\caption{Practical priors (E8) on Kuhn, $20$ seeds (mean $\pm$ sd). Expert (member-shaped) anchors steer
precisely; an unstructured risk tilt steers but is poorly predicted. Worst-case value is the game value
for all (exact equilibria).}
\label{tab:priors}
\begin{tabular}{lrrrr}
\toprule
reference & $\csel$ & I-proj.\ pred.\ & worst-case BR & median exploit. \\
\midrule
uniform             & $0.183\pm0.001$   & 0.216  & $-0.0564$  & $1\times10^{-3}$ \\
expert ($c{=}0.10$) & $0.107\pm0.000$   & 0.100  & $-0.0557$  & $4\times10^{-4}$ \\
risk-shaped         & $0.115\pm0.001$ & 0.029 & $-0.0558$ & $5\times10^{-4}$ \\
\bottomrule
\end{tabular}
\end{table}

\subsection{Two-dimensional steering}
\label{sec:2d}
\begin{figure}[t]
\centering
\includegraphics[width=0.55\textwidth]{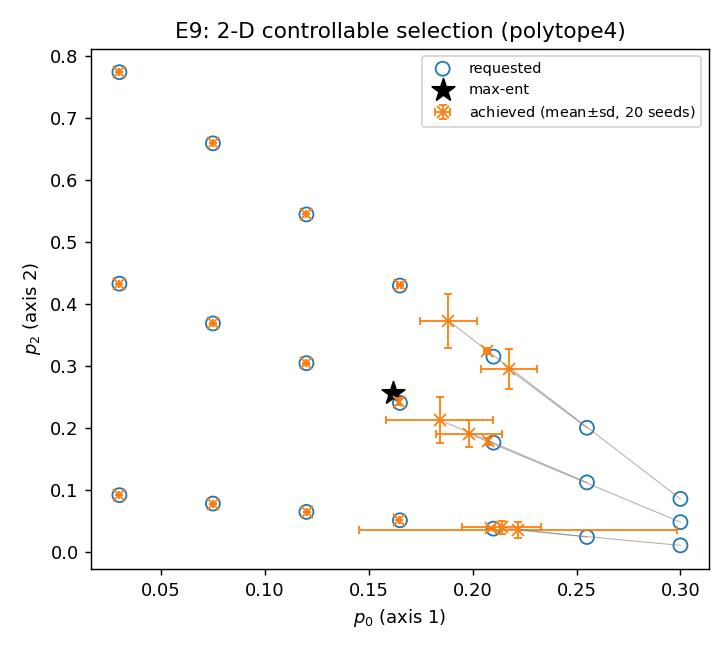}
\caption{Two-dimensional steering (E9) on polytope4: requested ($\circ$) vs achieved ($\times$, mean
$\pm$ sd over $20$ seeds) points in the $(p_0,p_2)$ plane; $\star$ is the maximum-entropy point.}
\label{fig:2d}
\end{figure}

On polytope4's genuine two-parameter Nash family ($20$ seeds; Fig.~\ref{fig:2d}), refined anchors steer
to a grid of $21$ 2-D targets with mean Euclidean error $0.042$ [0.040, 0.045]---but the
error distribution is sharply bimodal and diagnostic. The \emph{median} error is $0.0005$: interior
targets steer essentially exactly. The mean is carried by the six targets on the two rows nearest the
stiff-axis boundary ($p_0\in\{0.255,0.300\}$, mean error $0.139$, worst $0.308$ at a target whose
feasible $p_2$-range has shrunk to width $0.10$); the remaining $15$ targets steer with mean error
$0.0012$ (median $0.0002$, maximum $0.009$). Polytope4 contains the same skewed block as asym\_safe on
its $p_0$ coordinate, so this is a second, independent occurrence of the boundary-saturation effect of
Sec.~\ref{sec:control}, not a new failure mode. (Both are also a substantial improvement over the
single-seed demonstration of an earlier version, mean error $0.104$.) An off-manifold 2-D reference
lands near its I-projection prediction (predicted $(0.107, 0.172)$, measured
$(0.141, 0.175)\pm(0.001, 0.001)$; the residual on $p_0$ is the direct-readout ambiguity of
Sec.~\ref{sec:iproj}, as the reference is off-manifold), extending the approximate law to two
dimensions. The 2-D uniform baseline validates in Table~\ref{tab:val}.

\subsection{Steering follows the reference, not the initialization}
\label{sec:init}
All experiments above initialize policy parameters from the same small-scale Gaussian scheme,
independent of the reference. To rule out the residual worry that landing points reflect the
initialization basin rather than the reference, we cross them ($20$ seeds): initializing the policy
\emph{at} member($0.05$) while anchoring the reference at $0.30$ lands at
$0.300\pm0.000$; initializing at member($0.30$) while anchoring at $0.05$ lands at
$0.050\pm0.000$. In both cases the dynamics leave the initialization member and settle at the
reference anchor. Selection is governed by the reference, not by where the policy starts. This
complements the anchor-following finding of \citet{whichnash2026}, that selection moves with the
\emph{reference} initialization: jointly, the reference---and only the reference---is the controlling
object.
\section{Discussion}
\label{sec:discussion}
The results support a precise, bounded claim: \emph{the reference policy is a controllable steering wheel
for equilibrium selection in regularized self-play}, with the usable recipe being to anchor the reference
at the desired member and refine. This attains an essentially arbitrary polytope member to within
$\sim0.007$ at $\sim10^{-5}$--$10^{-4}$ exploitability across five games; the anchoring survives refinement;
and it is the reference, not the initialization, that does the selecting. The reach-weighted
I-projection is a good first-order model of where steering lands, exact in the on-manifold limit and
approximate at finite $\eta$.

We are equally explicit about the limits the data expose. Steering with a \emph{fixed off-manifold}
reference is not free: it realizes a regularized point whose exploitability scales with the reference's
distance from the polytope, so only the refinement recipe gives clean, near-Nash control. The solver's
mirror step must respect the Nash family's geometry: under prior-work defaults, stiff families fail to
converge and flat families yield near-Nash runs that scatter along the polytope---selection silently
fails while looking converged---which our baseline-only step rule detects and repairs. Even with the
right step, targets near the stiffest family's boundary undershoot: the dynamics settle on a genuine
but different member, so precision degrades exactly where the family's geometry is most extreme---and curvature predicts
where that saturation bites, while leaving interior precision untouched across a tenfold curvature
range. Proper testing also sharpens the architecture caveat: table--MLP
invariance is statistically established at every target, while the attention backbone's deviations,
though bounded by $0.04$, are real---significant at several targets and systematic in sign near the
family boundaries---so the correct statement, after the matched control arms, is: equivalence for table/MLP; for the
attention mechanism, robust excess variance with any systematic shift bounded at $\le0.018$ and
unestablished---which lands, with evidence, close to the variance-reading the earlier version asserted
without matched arms or equivalence tests, while bounding what that reading left open. Finally, the worst-case
selection--robustness trade-off is degenerate: because every polytope member is an exact equilibrium,
all share the game value against a best response, so steering changes only performance against fixed,
non-equilibrium opponents---a real but weaker effect than a worst-case trade-off.

\paragraph{Connection to reinforcement learning.} The mechanism we manipulate is the same
KL-to-reference term that appears throughout modern RL: the entropy bonus of maximum-entropy RL (uniform
reference), the trust region of TRPO/PPO (reference $=$ previous policy), and the KL-to-base-model
anchor of RLHF. That term is usually presented as a stabilizer. Our results recast it, in the
multi-agent equilibrium-selection setting, as a \emph{selection mechanism}: among many value-equivalent
solutions, the reference chooses which one is reached, predictably. The reference-steering recipe is, in
this light, the equilibrium-selection analogue of choosing an RLHF prior to select among
reward-equivalent policies.

\section{Limitations}
Our claims are confined to the exact-gradient, tractable-game regime: counterfactual values and best
responses are computed exactly on small games (2--12 information sets), with no sampling. The neural
architectures are miniature (a single-head attention block, a small MLP), stand-ins for backbone
\emph{mechanisms} rather than deep networks, and the architecture experiment is run on Kuhn only, under
the default solver configuration. Solver step sizes are per-game: four of the five 1-D games and polytope4
require a smaller mirror step for the dynamics to express the regularizer's selection at all, and
although the step was selected by a baseline-only rule and applied uniformly, sensitivity of selection
to the step size on stiff families is itself a finding (the boundary undershoot of
Sec.~\ref{sec:control}) and remains only partially characterized; we did not re-tune the neural
optimizers under the small step, so the architecture experiment is confined to Kuhn---a restriction
that is in any case forced: in our tokenizer every matrix-game information set emits a single token
(maxT $=1$, vs $4$ on Kuhn), and a self-attention block over one token collapses to a linear map with
no cross-information-set generalization pressure, so matrix games cannot exercise the attention
mechanism at all. Selection metrics use best-iterate reporting, a minimum-of-$N$ statistic that is
biased toward favorable iterates and more so for noisier arms; for between-arm contrasts this
attenuates rather than inflates differences, so the architecture deviations we report are, if
anything, conservative. The silent
scatter of near-Nash runs along flat families under too-large steps (Sec.~\ref{sec:val}) means
single-run selection claims should be distrusted in general unless validated against ground truth or
multiple seeds. With $n=5$ games, the curvature null result has power only against strong monotone laws. The
two-dimensional experiment uses $20$ seeds and one game. Finally, the I-projection law is verified as an
approximation at finite $\eta$, not proven; a finite-$\eta$ (QRE) correction and a formal statement of
when the approximation is tight are left to future work. The natural next tier---sampled best responses,
larger games, deeper backbones---is where the practical recipe and the boundary-stiffness caveat should
be stress-tested.

\section{Conclusion}
We asked whether the reference policy of regularized self-play can be used to choose the equilibrium on
purpose. It can: anchoring the reference at a target member and refining steers the dynamics to that
member at near-zero exploitability, predictably, independent of initialization, and---for the table
and MLP backbones---with statistically equivalent maps, with the selected member approximately the reach-weighted
I-projection of the reference. We have been deliberate about the boundaries of this result---the
exploitability cost of off-manifold references, the boundary undershoot on stiff families, the weak
identification of the uniform baseline on flat landscapes, the boundary-only curvature dependence, and the degeneracy
of the worst-case robustness trade-off---because the value of a controllable selection mechanism depends
on knowing exactly when it is clean. The same KL-to-reference knob is ubiquitous in reinforcement
learning; reading it as a selection mechanism, rather than only a stabilizer, is the conceptual
contribution we hope transfers beyond the tractable games studied here.

\section*{Reproducibility}
All games, equilibria, counterfactual values, exploitabilities, best responses, and the I-projection
oracle are computed exactly; the fast best-response oracle is validated bit-identical to pure-strategy
enumeration; the neural harness reproduces tabular R-NaD in the table limit. The full configuration
(per-game solver settings, seed counts of $20$ per game and $30$ per architecture arm, the
target/$\alpha$/$\eta$/refinement grids), the statistical protocol (bootstrap, TOST,
Kruskal--Wallis/Holm, Brown--Forsythe, Spearman), the complete source (written out as plain,
human-readable cells), and per-item checkpointing are provided in the accompanying notebook, which
regenerates every number, table, and figure in the Results sections from scratch on commodity
hardware in a few CPU-hours. The one exception is the step-size audit quoted in
Secs.~\ref{sec:protocol}--\ref{sec:val} (default-configuration failure exploitabilities and landing
scatter), which was run once while registering the per-game solver rule and is reported as a
diagnostic of that registration, not regenerated by the notebook.

\paragraph{Cross-build stability.} The reported dataset was produced on a cloud CPU runtime; the same
notebook was independently executed during development on a second machine with a different
NumPy/BLAS stack and identical seeds. Individual training runs are build-sensitive---per-run landed
coordinates differ by up to $0.023$ between builds, comparable to seed scatter, as expected for
chaotic iterative training under reordered floating-point reductions. Every reported statistic,
however, is distribution-level, and all of them reproduce across the two builds: across the $45$
architecture cells the largest per-arm mean shift is $0.001$ (an order below the per-arm standard
errors), the two builds select the same six Kruskal--Wallis-significant targets and the same TOST
equivalence fractions, and per-arm dispersions agree to three decimals. No conclusion in this paper
rests on a quantity that failed to reproduce across builds.

\appendix
\section{Configuration and metric definitions}
\label{app:repro}
\paragraph{Full configuration.} Seeds: $\{0,\dots,19\}$ for Kuhn table experiments (E2, E4, E5, E6, E8,
E10), $\{0,\dots,19\}$ for the four matrix games, $\{0,\dots,29\}$ for each arm of the architecture
experiment (including the matched arms), $\{0,\dots,19\}$ for polytope4. Architectures: table; MLP (hidden width $16$, $770$ parameters); single-head attention (hidden width $16$, head dimension $8$, $490$ parameters); matched arms: attention with head dimension $11$ ($787$ parameters) and MLP at Adam learning rate $0.012$. Solver: default games (kuhn, pennies\_safe) use $1200$ outer
iterations, mirror step $\lambda_q=1.0$, inner fitting steps $6$ (table) / $3$ (MLP, attention) at Adam
learning rates $0.05$/$0.015$/$0.012$; small-step games (asym\_safe, dup\_action, two\_safe, polytope4)
use $2000$ outer iterations, $\lambda_q=0.25$, $4$ inner steps at Adam learning rate $0.06$ (table
only; neural runs occur only on Kuhn). Default magnet $\eta=0.5$ with sweep
$\{2.0,1.5,1.0,0.7,0.5,0.35,0.25\}$; $9$ target coordinates per game placed at $3\%$--$97\%$ of the
coordinate range; mix ratios $\alpha\in\{0.15,0.3,0.45,0.6,0.75\}$ at $3$ member coordinates per game
($75$ E1 configurations); refinement periods $\{10,20,40,80,160\}$; best-iterate reporting with
snapshots every $\lfloor\text{iters}/60\rfloor$ iterations; gradient-norm clip $10$.

\paragraph{Metric definitions.} Exploitability is exact Nash-conv, computed by counterfactual
backward-induction best response (validated bit-identical to pure-strategy enumeration on $200$ random
strategy profiles per game). The selected coordinate is the family coordinate read directly from the
converged policy at the family-defining information set(s) (a 2-vector for polytope4); see
Appendix~\ref{app:oracle} for the readout-vs-projection check. Reach-weighted KL weights each
information set by its reach probability under the member being evaluated, normalized to sum to one.
Steering error is $|\csel-\ctar|$; the I-projection residual is
$|\csel-\argmin_c \KL_{\mathrm{reach}}(\mathrm{member}(c)\,\|\,\rho)|$. Selection metrics use
best-iterate exploitability.

\section{Oracle-definition checks}
\label{app:oracle}
Two definitional choices in the pipeline could in principle distort the readings; both are checked
explicitly.

\paragraph{Reach-weight normalization.} The I-projection oracle normalizes reach weights to sum to one,
which makes the objective a weighted average rather than the unnormalized sum
$\sum_I w_I(\sigma)\,\KL(\sigma_I\,\|\,\rho_I)$ written in some formulations; since the weights depend
on the member being evaluated, the two objectives can differ. On the full E1 reference grid ($15$
references per game), the argmin coordinate of the normalized and unnormalized objectives is identical
on the four matrix games and differs by at most 0.0050 ($1.5\%$ of the coordinate range, on the
flattest region of the Kuhn objective)---an order of magnitude below the E1 residual---so the oracle's
predictions are insensitive to this choice.

\paragraph{Coordinate readout vs projection.} The selected coordinate is read directly from the
family-defining information set of the converged policy. On the manifold this coincides with any
projection onto the family; off-manifold it need not. We therefore recompute the coordinate as a true
projection---the family member nearest in (i) unweighted $L_2$ distance over all information sets and
(ii) reach-weighted KL divergence---for every refined steering run (on-manifold) and every E1 run
(off-manifold, where exploitability reaches $0.08$--$0.25$). Direct readout and $L_2$ projection agree
to 0.0006 on average (max 0.0147) on-manifold, so all steering results are
readout-independent. Off-manifold the two conventions differ by 0.0162 on average (max 0.0740
at the most off-manifold references, $\alpha=0.75$; reach-KL projection differs from the readout by at
most 0.0637): an exploitable policy has no unique family coordinate, and this ambiguity is
\emph{comparable to the E1 residual itself}. Under the projection readout the I-projection law's mean
residual drops from 0.012 to 0.004 (Sec.~\ref{sec:iproj}); i.e.\ the law holds to a few
$10^{-3}$ once the coordinate is defined by projection, and the direct-readout residual should be read
as law error \emph{plus} readout ambiguity.

\end{document}